\documentclass{techreport}
\renewcommand{\reportshorttitle}{ARPS: Action-Relevant Predictive States}
\renewcommand{\reportdate}{\today}
\newcommand{\reportorganization}{DreamX Team}
\newcommand{\reportpdftitle}{The Right Future for Action: Learning Action-Relevant Predictive States in World Action Models}
\newcommand{\reportpdfauthor}{Qiwen Gu, Jifan Li, Bingjie Gao, Rui Chen, Jing Tang, Xiangxiang Chu, Junqiao Zhao}
\newcommand{\reportpdfsubject}{Action-Relevant Predictive States in World Action Models}
\newcommand{\reportpdfkeywords}{ARPS, world action models, predictive states, robot learning}
\setreportlogo{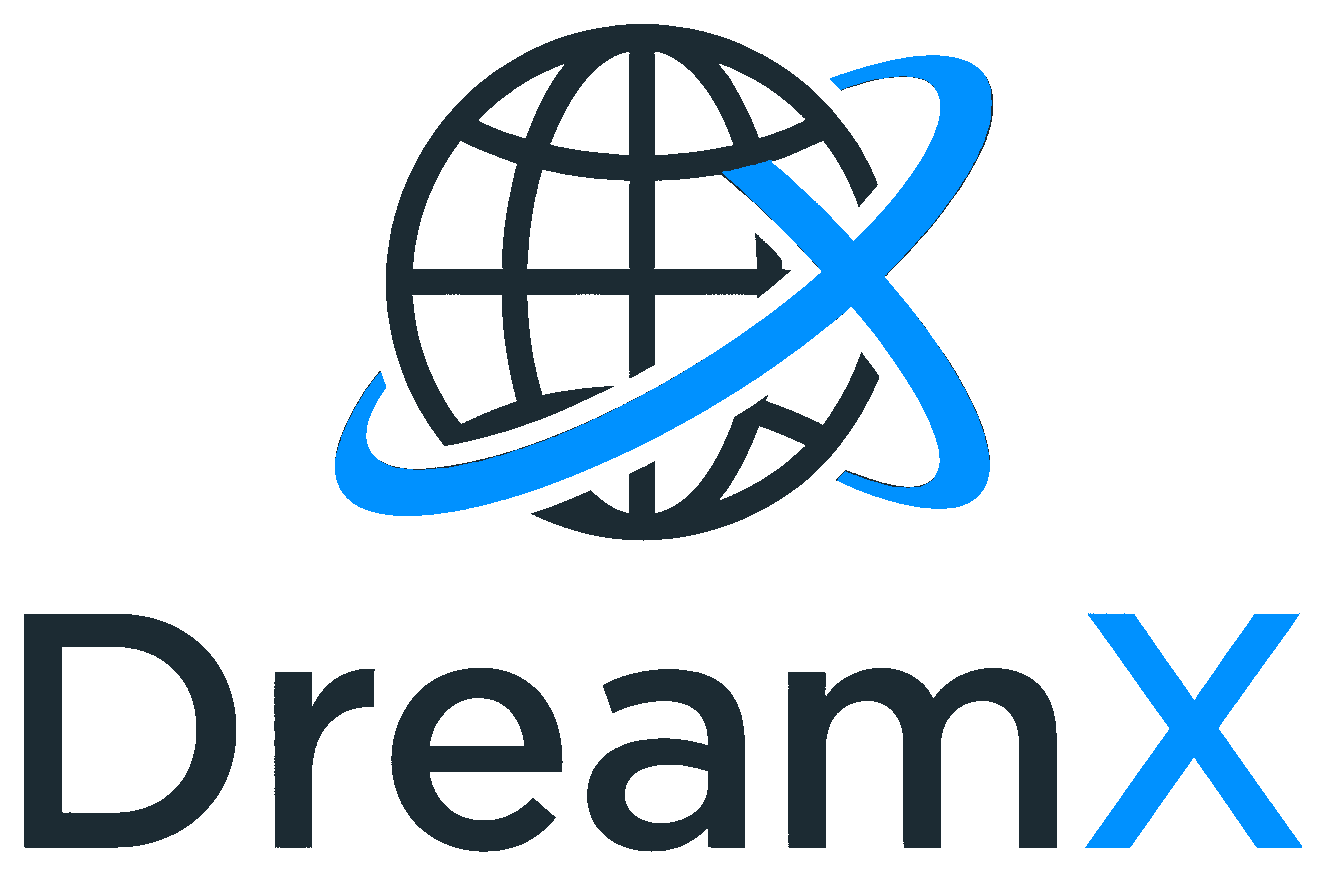}
\title{\reportpdftitle}
\author[1,2,\dagger]{Qiwen Gu}
\author[2,\ddagger]{Jifan Li}
\author[2,3,\dagger]{Bingjie Gao}
\author[2]{Rui Chen}
\author[2,\mathsection]{Jing Tang}
\author[2]{Xiangxiang Chu}
\author[1,\mathsection]{Junqiao Zhao}
\affiliation[1]{Tongji University}
\affiliation[2]{DreamX Team, Alibaba Group}
\affiliation[3]{Shanghai Jiao Tong University}
\abstract{Generation-free world action models (WAMs) retain future-video prediction during training but act from internal video features at inference, leaving unclear what these features should preserve for control.
Our representation diagnostics show that representations with more predictable future changes need not make linear action decoding easier. Observed future changes provide additional action information beyond the present, and linearly readable action information is spatially concentrated.
These findings motivate Action-Relevant Predictive States (ARPS), a compact predictive interface between the video and action experts.
ARPS uses a horizon-conditioned state predictor to aggregate intermediate video features into a compact state that supplies all visual context to the action expert.
Future-representation supervision trains different parts of this state to predict visual representations at different future times, together with their changes relative to the present.
At inference, the supervision branch is removed, and the action expert only uses the learned predictive state computed from current observations.
Controlled ablations show that future supervision substantially improves generalization under distribution shift.
ARPS achieves $99.2\%$ success on LIBERO and transfers to LIBERO-Plus without adaptation, reaching $87.3\%$ and exceeding Fast-WAM by $39.2$ percentage points.
}
\metadata[Status]{Technical Report}
\date{\reportdate}

\usepackage{amsmath}
\usepackage{amssymb}
\usepackage{array}
\usepackage{colortbl}
\usepackage{enumitem}
\usepackage{fancyhdr}
\usepackage{float}
\usepackage{xurl}

\definecolor{BrandPurple}{HTML}{7366CC}
\definecolor{BrandCyan}{HTML}{00B4E5}
\definecolor{LightGray}{HTML}{F5F5F5}
\definecolor{TableRowAlt}{HTML}{EAF4FB}
\definecolor{HeaderGray}{HTML}{888888}
\definecolor{TableRule}{HTML}{AEBBC5}

\newcommand{\reporttablehead}[1]{\textcolor{ReportText}{\textbf{#1}}}
\newcommand{\reporttablegroup}[1]{\textcolor{ReportPrimary}{\textbf{#1}}}
\newcommand{\reporttablepanelrow}{\rowcolor{ReportSurface}}

\newcommand{\reporttablehighlightrow}{\rowcolor{ReportSurface}}

\newcolumntype{L}[1]{>{\raggedright\arraybackslash}m{#1}}
\newcolumntype{C}[1]{>{\centering\arraybackslash}m{#1}}

\newtcolorbox{highlight}{
    colback=BrandCyan!5,
    colframe=BrandCyan,
    arc=2pt,
    boxrule=0.8pt,
    left=6pt, right=6pt, top=4pt, bottom=4pt,
    breakable
}

\newtcolorbox{keyfinding}{
    colback=BrandPurple!5,
    colframe=BrandPurple,
    coltitle=white,
    fonttitle=\bfseries,
    title=Key Finding,
    arc=2pt,
    boxrule=1pt,
    left=6pt, right=6pt, top=4pt, bottom=4pt,
    breakable
}

\newtcolorbox{tipbox}[1]{
    colback=ReportSurface,
    colframe=ReportPrimary,
    coltitle=white,
    fonttitle=\bfseries,
    title=#1,
    arc=2pt,
    boxrule=1pt,
    left=6pt, right=6pt, top=4pt, bottom=4pt,
    breakable
}

\fancypagestyle{plain}{%
    \fancyhf{}
    \fancyhead[L]{\footnotesize\color{HeaderGray}\sffamily \reportshorttitle}
    \fancyhead[R]{\footnotesize\color{HeaderGray} \thepage}
    \fancyfoot[C]{}

}

\fancypagestyle{firstpage}{%
    \fancyhf{}
    \fancyhead[L]{\small\sffamily\color{ReportPrimary} \reportorganization}
    \fancyhead[R]{\small\color{HeaderGray} \reportdate}
    \fancyfoot[C]{%
      \parbox{\textwidth}{%
        \raggedright\footnotesize
        $^{\dagger}$Work done during an internship at DreamX Team, Alibaba Group.\par
        $^{\ddagger}$Project lead. \par
        $^{\mathsection}$Corresponding authors. \par
        \vspace{3pt}\centering\color{gray}\thepage
      }%
    }

}

\usepackage{amsmath,amsfonts,bm}

\def\eqref#1{Eq.~\ref{#1}}

\def\1{\bm{1}}

\DeclareMathAlphabet{\mathsfit}{\encodingdefault}{\sfdefault}{m}{sl}
\SetMathAlphabet{\mathsfit}{bold}{\encodingdefault}{\sfdefault}{bx}{n}

\usepackage{multirow}
\usepackage{afterpage}

\hypersetup{
  pdftitle={\reportpdftitle},
  pdfauthor={\reportpdfauthor},
  pdfsubject={\reportpdfsubject},
  pdfkeywords={\reportpdfkeywords}
}
\begin{document}
\maketitle
\thispagestyle{firstpage}
\section{Introduction}
\label{sec:intro}

World action models (WAMs) learn to predict future visuals alongside generating actions, so video supervision can supply robot policies with information about dynamics and object interactions~\citep{DreamZero,LingBotVA,CosmosPolicy}.
Generation-free WAMs such as Fast-WAM retain this supervision during training but avoid costly future-video generation at inference, acting directly from internal video-expert features~\citep{FastWAM}.
These features form an implicit predictive state: they depend on the current observation but are shaped by learning what happens next.
What this state preserves for the action expert, however, remains unspecified.
Video prediction rewards appearance, geometry, semantics, and dynamics, while control may depend on only a subset of this information.
We ask: \emph{what should a predictive state preserve for action?}

The same future observation admits many valid representations, and predictive accuracy alone need not identify the one most useful for control.
Recent methods explore latent future prediction~\citep{LaWAM,LiLaWAM}, compact reasoning or world tokens~\citep{BeingH07,WorldTokens}, and representation-level supervision or alignment~\citep{Enfold,RepWAM,AGRA}.
These approaches demonstrate that useful foresight can be represented beyond pixels.
We examine which representation space, temporal content, and state capacity provide a useful interface between video modeling and action prediction.

Our diagnosis identifies three properties that guide this interface.
First, predictability can diverge from action readability: an easier-to-predict representation can expose less linearly decodable action structure.
Second, observed future changes add action-readable information beyond the present, with complementary gains across horizons.
Third, action readability is spatially concentrated, suggesting that the action interface need not retain the full dense representation.
These findings motivate a teacher selected for action readability, multi-horizon absolute and temporal-difference targets, and a learned capacity bottleneck.

We introduce \textbf{Action-Relevant Predictive States (ARPS)} to implement these choices.
Horizon-conditioned learnable queries aggregate intermediate video-expert features into a compact state that supplies all visual context to the action expert.
Training-only heads align this state with absolute and temporal-difference targets from a frozen V-JEPA2 teacher, alongside the original video and action objectives.
At inference, the teacher and decoding heads are removed, and the video expert terminates at the state-extraction layer.

ARPS achieves $99.2\%$ on LIBERO and $87.3\%$ on LIBERO-Plus without adaptation, exceeding Fast-WAM by $39.2$ points on the latter and reducing end-to-end latency by $13\%$.
On RoboTwin 2.0, ARPS retains the baseline's overall dual-arm manipulation performance, with $91.32\%$ on Clean and $90.10\%$ on Random.
Ablations show that future supervision substantially improves OOD performance, with gains from a farther horizon and joint absolute and temporal-difference targets.
State analyses link the learned geometry to future behavior and nuisance invariance, while interventions verify its use in closed-loop control.

Our main contributions are:
\begin{itemize}
    \item We diagnose the content of predictive representations for control: predictability and action readability can diverge, observed future changes add complementary action signal, and action readability is spatially concentrated.
    \item We introduce \textbf{ARPS}, which uses these findings to construct compact, horizon-conditioned states from intermediate WAM features with absolute and temporal-difference representation supervision.
    \item We validate the state design through ablations, representation analyses, and closed-loop interventions. ARPS improves in-domain and out-of-distribution control on LIBERO and LIBERO-Plus, maintains overall RoboTwin performance, and reduces inference latency.
\end{itemize}

\section{Related Work}
\label{sec:related}

\textbf{Pretrained representations and vision-language-action policies.}
Pretrained visual encoders provide different representational priors: DINOv2 emphasizes discriminative features, V-JEPA2 learns via latent video prediction, and VGGT encodes dense geometric structure~\citep{DINOv2,VJEPA2,VGGT}.
These complementary objectives motivate our comparison of representation spaces for action prediction.
VLAs adapt vision-language backbones for robot control, with progress in generalist manipulation, continuous action generation, and generalization through heterogeneous co-training~\citep{OpenVLA,PiZero,PiZeroPointFive}.
Recent work further addresses cross-embodiment transfer and efficient action modeling~\citep{XVLA,ABotM0}, while predictive representation learning connects VLAs to latent world modeling~\citep{VLAJEPA}.

\textbf{World action models and representation-level foresight.}
World action models couple future visual prediction with action generation to
transfer physical and interaction structure from pretrained video and world
models~\citep{PhyParam,DreamXWorld}.
DreamZero and LingBot-VA jointly model video and action, Motus unifies
understanding, video generation, and actions, and Cosmos Policy adds
future-state and value prediction, while DreamX-Phi conditions a video world
model on actions for robotic manipulation~\citep{DreamZero,LingBotVA,Motus,CosmosPolicy,DreamXPhi}.
To avoid visual imagination, other methods act through generation-free or
latent interfaces~\citep{FastWAM,LaWAM,LiLaWAM,BeingH07,WorldTokens}.
Enfold distills multi-level future-conditioned states into a compact current-only
representation, whereas DC-WAM redirects supervision and attention from appearance
toward temporal changes and interaction regions~\citep{Enfold,DCWAM}.
Representation-space foresight is also used directly: VPP conditions inverse
dynamics on predicted features~\citep{VPP}, while RepWAM and AGRA align world-model features for control~\citep{RepWAM,AGRA}.
ARPS first tests which future representations and changes are action-readable,
then compresses that information into horizon-conditioned states.

\textbf{Compact action-facing interfaces.}
Attention bottlenecks convert dense perception into fixed-capacity tokens:
Perceiver, Slot Attention, and TokenLearner establish query-based aggregation,
object-centric abstraction, and adaptive token selection, while PerAct and RT-1
bring these mechanisms to robot control~\citep{Perceiver,SlotAttention,TokenLearner,PerAct,RT1}.
Recent work makes compression control- or dynamics-aware: Compressor-VLA uses
instruction-conditioned visual queries, Token Bottleneck compresses scene dynamics
into one token, and ThinkAct and Fast-ThinkAct encode embodied reasoning as compact
visual or latent plans for downstream action~\citep{CompressorVLA,TokenBottleneck,ThinkAct,FastThinkAct}.
ARPS likewise uses learnable queries, but multi-horizon future-representation
supervision determines what predictive content passes through its bottleneck to the action expert.

\section{What Should a Predictive State Preserve?}
\label{sec:diagnosis}

\begin{figure}[!t]
    \centering
    \includegraphics[width=\linewidth]{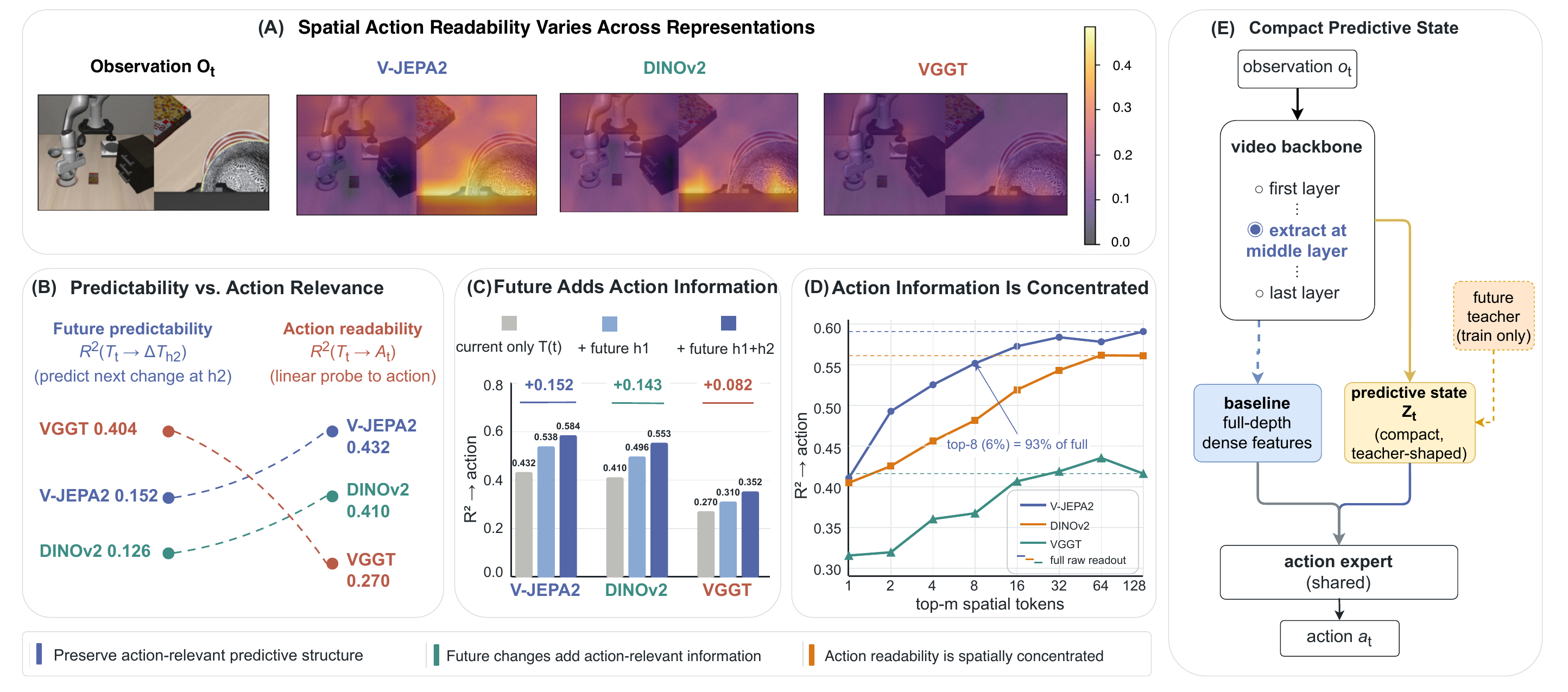}
\caption{Predictive-state diagnosis. The probes compare future-change predictability with action readability, measure the action information added by observed futures, and test how quickly readability saturates across dense spatial features. The rightmost schematic previews the compact predictive state motivated by these findings.}
    \label{fig:diagnosis}
\end{figure}

Generation-free WAMs remove test-time future-video denoising, but leave unclear what their action-facing video features preserve.
We examine Fast-WAM~\citep{FastWAM}, whose separate video and action DiTs interact through masked shared attention.
Let $A_t$ denote the action chunk and $H_t=F_\theta(O_t)$ the current-frame features exposed to the action expert.
Its training objective is
\begin{equation}
    \mathcal{L}_{\mathrm{WAM}}
    =
    \mathcal{L}_{\mathrm{video}}
    \bigl(O_{t:t+H},\theta\bigr)
    +
    \lambda_{\mathrm{act}}
    \mathcal{L}_{\mathrm{act}}
    \bigl(A_t\mid H_t,\phi\bigr).
    \label{eq:fastwam_training}
\end{equation}
Here, $O_{t:t+H}$ includes the current and future latents. The mask hides future latents from $H_t$, although their prediction still shapes $F_\theta$.
At inference, future tokens are omitted and the action expert predicts $\hat{A}_t=\pi_\phi(H_t)$.
This makes $H_t$ an implicit predictive state: it is conditioned on the present but shaped by future prediction during training.
The objective does not guarantee that the state preserves action-relevant future information, nor does it establish that a dense state is necessary.
We examine the choice of representation space, the action information supplied by observed futures, and the amount of dense information needed for readout.
The answers inform the teacher, predictive objective, and capacity used in ARPS.
These diagnostics compare candidate teacher spaces for supervising the state, rather than directly measuring the internal WAM features $H_t$.

\textbf{Diagnostic protocol.}
We analyze 2,048 samples from the LIBERO training set with a 60/20/20\% episode-level split, pairing each current observation and its future with a 32-step action chunk.
Linear ridge probes report test $R^2$ relative to a mean-prediction baseline.
Ridge regularization and spatial ranking are selected only on validation data.
App.~\ref{app:diagnostics} gives the full probe protocol.

\subsection{Predictability Does Not Imply Action Readability}
\label{sec:action_relevance}

We begin by asking whether an easier-to-predict representation also exposes more information to a simple action readout.
We compare three frozen encoders with distinct pretraining biases: V-JEPA2~\citep{VJEPA2} toward temporal prediction, DINOv2~\citep{DINOv2} toward semantic discrimination, and VGGT~\citep{VGGT} toward geometry.
For each encoder $T$, we measure linear action readability by probing from the current representation $T_t$ to the action chunk $A_t$.

Fig.~\ref{fig:diagnosis}(a) visualizes held-out action-probe $R^2$ from each spatial location individually.
The markedly different patterns show that the encoders expose different linearly action-readable structure for the same observation.
Their spatially pooled probes achieve $R^2=0.432$, $0.410$, and $0.270$ for V-JEPA2, DINOv2, and VGGT, respectively (Fig.~\ref{fig:diagnosis}(b)).

To compare action readability with predictability, we define the future representation change as
$
\Delta T_h = T_{t+h}-T_t
$
and fit a separate probe from $T_t$ to $\Delta T_{h_2}$.
VGGT is most predictable, with $R^2=0.404$ versus $0.152$ for V-JEPA2 and $0.126$ for DINOv2, yet it is least action-readable.
The mismatch appears under the same observations and evaluation protocol.
Ease of predicting future change is therefore not a reliable proxy for action readability, much less downstream control utility.
We use V-JEPA2 as the default teacher because it provides the strongest linear action readout.

\subsection{Future Information Adds Beyond the Present}
\label{sec:future_increment}

The previous analysis measures present-time readability but not the additional information in observed futures.
We form $T^{+}=[T_t,\Delta T_{h_1},\Delta T_{h_2}]$ and compare probes using $T_t$, $[T_t,\Delta T_{h_1}]$, and $T^{+}$ under the same protocol.
This is an oracle-information diagnostic rather than an inference-time construction.
Observed future frames reveal a useful supervision signal, while ARPS must preserve that information in a state computed from the current observation alone.
Fig.~\ref{fig:diagnosis}(c) shows consistent gains across all three representations.
V-JEPA2 improves from $0.432$ to $0.538$ and $0.584$ after adding one and two future changes.
DINOv2 improves from $0.410$ to $0.496$ and $0.553$, while VGGT improves from $0.270$ to $0.310$ and $0.352$.
The corresponding two-horizon gains over present-only input are $0.152$, $0.143$, and $0.082$.

Observed future changes contain action information not linearly exposed by the present representation, and the second horizon adds complementary signal.
This diagnostic motivates temporal-difference supervision at multiple horizons.
We pair these targets with absolute future representations to retain scene context, and test the complementary contribution of the two objectives through the control ablations in Sec.~\ref{sec:ablation}.

\afterpage{%
\begin{figure}[!t]
    \centering
    \includegraphics[width=\linewidth]{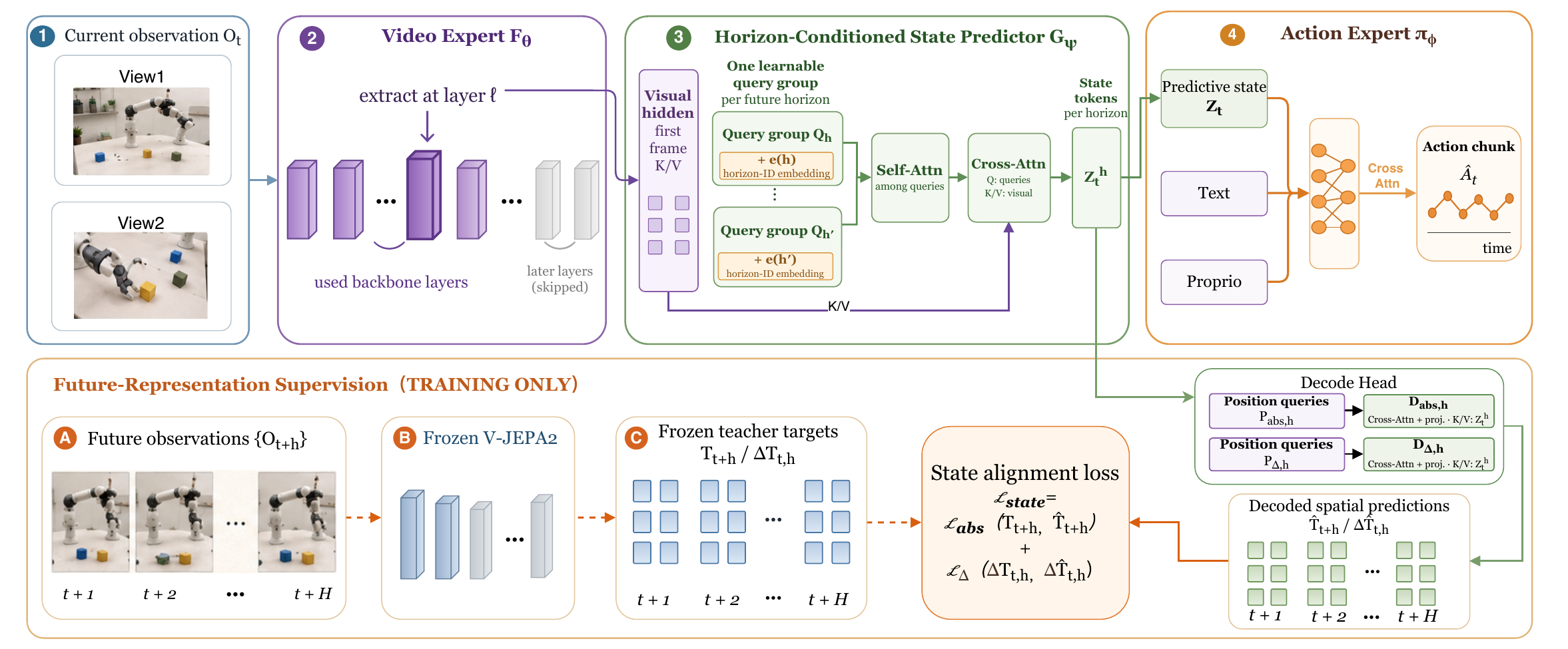}
    \caption{Overview of ARPS. The horizon-conditioned state predictor turns intermediate video features into the action expert's predictive state. Training-only future-representation supervision aligns decoded predictions with absolute and temporal-difference targets from frozen V-JEPA2. The teacher and decoding heads are removed at inference.}
    \label{fig:overview}
\end{figure}
}
\vspace{-5pt}
\subsection{Action Readability Is Spatially Concentrated}
\label{sec:selective_compression}

We next ask whether the action policy must consume the entire dense spatial representation.
For each teacher, we center every current-frame feature $T_t^p$ by its training mean and rank the 128 locations by validation ridge-probe $R^2$ to the action chunk.
For $m \in \{1,2,4,8,16,32,64,128\}$, we flatten the top-$m$ features, refit the probe, and report held-out action $R^2$.
The full 128-location readout defines the reference ($R^2=0.5907$ for V-JEPA2).
Fixing the ranking on validation data makes this an oracle concentration diagnostic without test-set selection.
Fig.~\ref{fig:diagnosis}(d) shows that action readability saturates well before all locations are retained.
Eight V-JEPA2 locations reach $R^2=0.551$, recovering $93.3\%$ of the full readout from only $6.25\%$ of the field.
DINOv2 and VGGT show the same trend.
Dense visual representations are spatially overcomplete for linear action readout.

This concentration motivates a learned capacity bottleneck, but does not prescribe a particular compression architecture or imply that a fixed subset suffices for control.
Because informative locations may vary with the scene and task, ARPS aggregates content-dependent evidence into a compact state rather than hard-coding a spatial selection rule.

\textbf{Design implication.}
ARPS aggregates dense WAM features into a compact learned state and supervises it with multi-horizon absolute and temporal-difference targets in the action-readable V-JEPA2 space.
Its teacher, objective, and capacity each follow from a corresponding diagnostic above.

\section{Method}
\label{sec:method}

Sec.~\ref{sec:diagnosis} motivates three design choices: an action-readable V-JEPA2 teacher, absolute-and-change supervision over multiple horizons, and a learned capacity bottleneck.
\textbf{Action-Relevant Predictive States (ARPS)} combines a horizon-conditioned state predictor with future-representation supervision, as illustrated in Fig.~\ref{fig:overview}.
All visual context supplied to the action expert passes through this state, while the original video objective continues to train the world model.

\subsection{Overview}
\label{sec:method_overview}

Let $O_t$ denote the current multi-view observation and let $F_\theta$ be the video expert.
ARPS reads the first-frame hidden tokens after an intermediate layer $\ell$, denoted by $H_t^\ell=F_\theta^{1:\ell}(O_t)$.
The horizon-conditioned state predictor $G_\psi$ converts $H_t^\ell$ into the predictive state $Z_t$, from which the action expert predicts $\hat A_t=\pi_\phi(Z_t)$.
We omit text and proprioception from this notation.
The first-frame causal mask prevents $H_t^\ell$, and hence $Z_t$, from accessing future video tokens.
During training, the full video expert remains active for the video loss, while the action-facing path reads only $H_t^\ell$.
At inference, video-expert computation stops at layer $\ell$ and the complete state is supplied to the action expert.

\subsection{Horizon-Conditioned State Predictor}
\label{sec:state_bottleneck}

The spatial concentration observed in Sec.~\ref{sec:selective_compression} motivates replacing dense WAM features with a fixed-capacity learned state.
The predictor initializes $M$ learnable state queries and divides them evenly among the supervised future horizons $\mathcal{H}$.
For each $h\in\mathcal{H}$, $Q_h$ is a distinct query group assigned to that horizon, with a horizon embedding $e_h$ added to the group.
The complete predictive state is defined as
\begin{equation}
    Z_t
    =
    G_\psi\!\left(
        H_t^\ell,
        \mathop{\operatorname{concat}}_{h\in\mathcal{H}}(Q_h+e_h)
    \right)
    =
    \mathop{\operatorname{concat}}_{h\in\mathcal{H}} Z_t^{(h)},
    \label{eq:state_predictor}
\end{equation}
where $e_h$ is broadcast across its group and $Z_t^{(h)}$ contains the observation-dependent states produced from the queries assigned to horizon $h$.

The horizon groups specify parameter initialization and supervision routing, but they do not isolate information flow.
Within every predictor block, unmasked self-attention operates jointly over all $M$ state tokens, allowing different horizon groups to exchange information.
The resulting tokens then cross-attend to $H_t^\ell$, which supplies keys and values.
The heads for horizon $h$ read $Z_t^{(h)}$, a subset of the shared state.
This globally coupled but horizon-labeled construction lets the compact state share evidence across future times while preserving an explicit supervision route for each one.

\subsection{Future-Representation Supervision}
\label{sec:future_supervision}

The predictor fixes the capacity of $Z_t$, while a frozen V-JEPA2 teacher shapes its content with future-representation targets, as shown in Fig.~\ref{fig:overview}.
From ground-truth training trajectories, the teacher extracts a current spatial field $T_t$ and a future field $T_{t+h}$ for each $h\in\mathcal H$.
Here, $h$ indexes future video latents, and each future field averages teacher features over the sampled frames represented by that latent.
We define $\Delta T_{t,h}=T_{t+h}-T_t$ and detach all targets.
App.~\ref{app:implementation} details the causal clip construction, temporal pooling, and per-channel standardization.

The state and teacher fields generally contain different numbers of tokens.
Position-query heads predict dense teacher fields from the compact state, providing spatially detailed supervision without supplying teacher features to the action expert.
For each horizon, ARPS uses an absolute and a temporal-difference decoding head,
\begin{equation}
    \hat{T}_{t+h}
    =D_{\mathrm{abs},h}\bigl(Z_t^{(h)}\bigr),
    \qquad
    \widehat{\Delta T}_{t,h}
    =D_{\Delta,h}\bigl(Z_t^{(h)}\bigr),
    \label{eq:state_decoders}
\end{equation}
where $D_{\mathrm{abs},h}$ predicts the absolute future representation and $D_{\Delta,h}$ predicts its change from the current representation.
Each head uses its own learned position queries, one per teacher spatial location.
The position queries cross-attend to $Z_t^{(h)}$, which supplies keys and values, and the outputs are projected into the teacher space.
The heads have separate parameters for each target and horizon.
The targets play complementary roles.
The absolute field $T_{t+h}$ retains future scene and object context, whereas $\Delta T_{t,h}$ removes the current baseline and emphasizes the temporal variation identified in Sec.~\ref{sec:future_increment}.

We match a prediction $\hat X$ to its teacher target $X$ using equally weighted cosine and smooth-$\ell_1$ terms,
\begin{equation}
    \ell_{\mathrm{rep}}(\hat X,X)
    =1-\operatorname{cos}(\hat X,X)
    +\operatorname{SmoothL1}(\hat X,X),
    \label{eq:representation_matching}
\end{equation}
where smooth-$\ell_1$ uses its default transition parameter $\beta=1$.
The loss is evaluated at each teacher spatial position and averaged over positions.
We suppress that index below.
We jointly supervise the absolute and temporal-difference predictions as
\begin{equation}
    \mathcal L_{\mathrm{state}}
    =\mathcal L_{\mathrm{abs}}+\mathcal L_{\Delta}
    =
    \frac{1}{|\mathcal{H}|}\sum_{h\in\mathcal{H}}
    \left[
        \lambda_{\mathrm{state}}\,
        \ell_{\mathrm{rep}}(\hat T_{t+h},T_{t+h})
        +\lambda_{\Delta}\,
        \ell_{\mathrm{rep}}(\widehat{\Delta T}_{t,h},\Delta T_{t,h})
    \right].
    \label{eq:future_rep_loss}
\end{equation}
Joint supervision therefore asks the compact state to retain both the future representation and how it changes from the present.
Here, $\mathcal L_{\mathrm{abs}}$ and $\mathcal L_{\Delta}$ denote the weighted absolute and temporal-difference losses, each averaged over horizons.
The heads read only $Z_t^{(h)}$. Teacher features provide loss targets but never enter the action-facing forward path.

\subsection{Predictive-State-Conditioned Action Prediction}
\label{sec:state_conditioning}

The action expert consumes the complete, globally coupled state rather than separate horizon groups.
ARPS projects $Z_t$ once, appends it to the existing text and proprioceptive context, and reuses the resulting context as keys and values in every action-expert block.
The dense WAM features are not appended separately, so all visual information available to the action expert must pass through $Z_t$.

Unlike the Fast-WAM-style objective in Eq.~\ref{eq:fastwam_training}, whose action term reads the implicit representation $H_t$, ARPS predicts actions from the explicit state $Z_t$ and optimizes
\begin{equation}
    \mathcal L_{\mathrm{ARPS}}
    =
    \mathcal L_{\mathrm{video}}
    \bigl(O_{t:t+H},\theta\bigr)
    +
    \lambda_{\mathrm{act}}
    \mathcal L_{\mathrm{act}}
    \bigl(A_t\mid Z_t\bigr)
    +
    \mathcal L_{\mathrm{state}}.
    \label{eq:arps_objective}
\end{equation}

At inference, the visual pathway reduces to
$
O_t \rightarrow H_t^\ell \rightarrow Z_t \rightarrow \hat{A}_t
$.
The frozen teacher and all horizon- and target-specific decoding heads are removed.
Only video-expert layers $1,\ldots,\ell$ are evaluated before the predictor forms $Z_t$ and the action expert predicts actions.

\section{Experiments}
\label{sec:experiments}


\textbf{Setup.} We build ARPS on Fast-WAM~\citep{FastWAM}, using Wan2.2-TI2V-5B as the video expert and a 1B ActionDiT as the action expert, with ActionDiT initialized from the video expert.
Unless stated otherwise, we extract layer-15 features and use a three-block predictor with $M=32$ states split evenly across the two future latent horizons $\mathcal H=\{1,2\}$, supervised by a frozen V-JEPA2 teacher.
We set $\lambda_{\mathrm{act}}=\lambda_{\mathrm{state}}=\lambda_{\Delta}=1.0$ throughout.
We evaluate ARPS on three benchmarks: LIBERO~\citep{LIBERO},
LIBERO-Plus~\citep{LIBEROPlus}, and RoboTwin 2.0~\citep{RoboTwin2}.
Following the LIBERO-Plus protocol, we directly evaluate the policy trained on LIBERO without adaptation.
Additional implementation details are provided in App.~\ref{app:implementation}.

\subsection{Main Results}
\label{sec:libero_results}

We compare ARPS with representative VLAs and WAMs~\citep{OpenVLA,PiZero,PiZeroPointFive,ABotM0,XVLA,VLAJEPA,Motus,GigaWorldPolicy,BeingH07,LiLaWAM,LingBotVA,Enfold,DCWAM,FastWAM,CosmosPolicy,FasterWAM}.
LIBERO tests in-domain control, RoboTwin 2.0 tests applicability to dual-arm manipulation, and LIBERO-Plus evaluates the same LIBERO-trained policy under distribution shift.
Fast-WAM results on RoboTwin 2.0 and LIBERO-Plus are from our reproduction runs.

\begin{table}[t]
    \centering
    \begingroup
    \caption{LIBERO success rates (\%). ``Embodied PT.'' indicates whether embodied pretraining is used.}
    \label{tab:libero_results}
    \small
    \setlength{\tabcolsep}{7pt}
    \renewcommand{\arraystretch}{1.15}
    \arrayrulecolor{TableRule}
    \begin{tabular}{@{}clcccccc@{}}
        \toprule
        \reporttablepanelrow
        \reporttablehead{Family} & \reporttablehead{Method} & \reporttablehead{Embodied PT.} & \reporttablehead{Spatial} & \reporttablehead{Object} & \reporttablehead{Goal} & \reporttablehead{Long} & \reporttablehead{Average} \\
        \midrule
        \multirow{4}{*}{VLA} & OpenVLA
            & $\checkmark$ & 84.7 & 88.4 & 79.2 & 53.7 & 76.5 \\
        & $\pi_0$
            & $\checkmark$ & 96.8 & 98.8 & 95.8 & 85.2 & 94.1 \\
        & $\pi_{0.5}$
            & $\checkmark$ & \underline{98.8} & 98.2 & \underline{98.0} & 92.4 & 96.9 \\
        & VLA-JEPA
            & $\checkmark$ & 96.2 & 99.6 & 97.2 & 95.8 & 97.2 \\
        \cmidrule(lr){1-8}
        \multirow{6}{*}{WAM} & Motus
            & $\checkmark$ & 96.8 & \underline{99.8} & 96.6 & 97.6 & 97.7 \\
        & LingBot-VA
            & $\checkmark$ & 98.5 & 99.6 & 97.2 & \underline{98.5} & \underline{98.5} \\
        & Enfold
            & $\times$ & 97.0 & \textbf{100.0} & 96.8 & 97.4 & 97.8 \\
        & DC-WAM
            & $\times$ & -- & -- & -- & -- & 98.1 \\
        & Fast-WAM
            & $\times$ & 98.2 & \textbf{100.0} & 97.0 & 95.2 & 97.6 \\
        \cmidrule(lr){2-8}
        \reporttablehighlightrow
        & \textbf{ARPS (Ours)}
            & $\times$ & \textbf{99.0} & \textbf{100.0} & \textbf{98.6} & \textbf{99.0} & \textbf{99.2} \\
        \bottomrule
    \end{tabular}
    \arrayrulecolor{black}
    \endgroup
\end{table}

\textbf{LIBERO.}
ARPS achieves the highest average among the methods in Tab.~\ref{tab:libero_results} at $99.2\%$, exceeding LingBot-VA by $0.7$ points without embodied pretraining.
It improves over Fast-WAM by $1.6$ points overall and $3.8$ on LIBERO-Long.
These results support the combined design of a learned state interface and teacher-guided predictive supervision: all visual context passes through 32 state tokens while retaining strong control, including on the long-horizon suite.
The nearly saturated scores, however, offer limited resolution for comparing state designs.

\begin{table}[t]
    \centering
    \begingroup
    \caption{RoboTwin 2.0 success rates (\%).}
    \label{tab:robotwin_results}
    \small
    \setlength{\tabcolsep}{8pt}
    \renewcommand{\arraystretch}{1.15}
    \arrayrulecolor{TableRule}
    \begin{tabular}{@{}clccc@{}}
        \toprule
        \reporttablepanelrow
        \reporttablehead{Family} & \reporttablehead{Method} & \reporttablehead{Clean} & \reporttablehead{Random} & \reporttablehead{Average} \\
        \midrule
        \multirow{4}{*}{VLA} & $\pi_0$ & 65.92 & 58.40 & 62.2 \\
        & $\pi_{0.5}$ & 82.74 & 76.76 & 79.8 \\
        & ABot-M0 & 81.20 & 80.40 & 80.8 \\
        & X-VLA & 72.88 & 72.84 & 72.9 \\
        \cmidrule(lr){1-5}
        \multirow{6}{*}{WAM} & Motus & 88.66 & 87.02 & 87.8 \\
        & GigaWorld-Policy & 86.36 & 85.04 & 85.7 \\
        & Being-H0.7 & 90.20 & 89.60 & \underline{89.9} \\
        & LiLa-WAM & \underline{90.48} & 89.04 & 89.8 \\
        & Fast-WAM & 90.42 & \textbf{91.06} & \textbf{90.7} \\
        \cmidrule(lr){2-5}
        \reporttablehighlightrow
        & \textbf{ARPS (Ours)} & \textbf{91.32} & \underline{90.10} & \textbf{90.7} \\
        \bottomrule
    \end{tabular}
    \arrayrulecolor{black}
    \endgroup
\end{table}

\textbf{RoboTwin 2.0.}
Among the methods in Tab.~\ref{tab:robotwin_results}, ARPS has the highest Clean success rate at $91.32\%$ and an average of $90.7\%$.
Compared with Fast-WAM, it gains $0.9$ points on Clean and loses $1.0$ on Random, yielding comparable overall performance.
The teacher-supervised predictive state thus remains effective for dual-arm manipulation, although its advantage depends on the evaluation setting.

\begin{table}[t]
    \centering
    \begingroup
    \caption{
    Success rates (\%) on LIBERO-Plus under different distribution shifts.
    }
    \label{tab:libero_plus}
    \small
    \setlength{\tabcolsep}{4.0pt}
    \renewcommand{\arraystretch}{1.15}
    \arrayrulecolor{TableRule}
    \begin{tabular}{clcccccccc}
        \toprule
        \reporttablepanelrow
        \reporttablehead{Family} & \reporttablehead{Method}
        & \reporttablehead{Camera} & \reporttablehead{Robot} & \reporttablehead{Language} & \reporttablehead{Light}
        & \reporttablehead{Background} & \reporttablehead{Noise} & \reporttablehead{Layout} & \reporttablehead{Total} \\
        \midrule
        \multirow{4}{*}{VLA} & OpenVLA
        & 0.8 & 3.5 & 23.0 & 8.1
        & 34.8 & 15.2 & 28.5 & 15.6 \\

        & $\pi_0$
        & 13.8 & 6.0 & 58.8 & 85.0
        & 81.4 & 79.0 & 68.9 & 53.6 \\

        & $\pi_{0.5}$
        & 64.8 & \textbf{71.8} & 83.0 & 93.5
        & \underline{92.2} & 78.8 & \textbf{85.5} & 81.4 \\

        & VLA-JEPA
        & 63.3 & 67.1 & 85.4 & 95.6
        & \textbf{93.6} & 66.3 & \underline{85.1} & 79.5 \\
        \cmidrule(lr){1-10}
        \multirow{6}{*}{WAM} & Fast-WAM
        & 14.3 & 41.0 & 65.6 & 77.8
        & 51.8 & 37.0 & 60.0 & 48.1 \\

        & Faster-WAM
        & 53.8 & \underline{71.6} & \underline{94.7} & 96.3
        & 61.3 & 63.6 & 79.1 & 73.6 \\

        & Enfold
        & 59.2 & 61.0 & 92.1 & 89.7
        & 48.7 & 35.8 & 76.6 & 66.2 \\

        & DC-WAM
        & 23.9 & 51.7 & 83.4 & 91.7
        & 61.3 & 54.2 & 69.8 & 60.9 \\

        & Cosmos-Policy
        & \underline{75.8} & 63.3 & 81.7 & \underline{96.5}
        & 88.9 & \underline{92.7} & 82.2 & \underline{82.2} \\

        \reporttablehighlightrow
        & \textbf{ARPS (Ours)}
        & \textbf{88.6} & 62.8 & \textbf{97.8} & \textbf{97.5}
        & 91.5 & \textbf{95.0} & 81.8 & \textbf{87.3} \\

        \bottomrule
    \end{tabular}
    \arrayrulecolor{black}
    \endgroup
\end{table}

\textbf{LIBERO-Plus.}
Without adaptation, ARPS reaches $87.3\%$, exceeding Fast-WAM by $39.2$ points and the strongest baseline, Cosmos-Policy, by $5.1$ (Tab.~\ref{tab:libero_plus}).
The much larger gap to Fast-WAM than on LIBERO shows that similar in-domain success can conceal substantial differences under shift.
Gains are strongest for Camera ($74.3$ points) and Noise ($58.0$), although $\pi_{0.5}$ remains better on Robot and Layout, and VLA-JEPA on Background.
ARPS improves over its Fast-WAM backbone in all seven categories, so the overall gain is not confined to a single perturbation type.
Together with the LIBERO results, this shows that the compact state can retain in-domain competence while improving robustness under shift, without dominating every category across all methods.
These results establish the effectiveness of the complete state design.

\subsection{Ablation Study}
\label{sec:ablation}

Tab.~\ref{tab:arps_ablation} tests the representation, temporal targets, and capacity motivated by Sec.~\ref{sec:diagnosis}, together with extraction depth and gradient flow.
Each variant changes one component of the default configuration.
We report both LIBERO and LIBERO-Plus to distinguish in-domain control from robustness under distribution shift.
App.~\ref{app:full_libero_ablation} provides the complete per-suite LIBERO results and additional objective and interface comparisons.

\begin{table}[t]
    \centering
    \begingroup
    \caption{ARPS ablations on LIBERO and LIBERO-Plus. LIBERO reports only the average success rate. }
    \label{tab:arps_ablation}
    \footnotesize
    \setlength{\tabcolsep}{1.35pt}
    \renewcommand{\arraystretch}{1.15}
    \arrayrulecolor{TableRule}
    \begin{tabular}{@{}llccccccccccc@{}}
        \toprule
        \reporttablepanelrow
        & & \multicolumn{2}{c}{\reporttablegroup{LIBERO}}
        & \multicolumn{9}{c}{\reporttablegroup{LIBERO-Plus}} \\
        \cmidrule(lr){3-4}\cmidrule(lr){5-13}
        \reporttablepanelrow
        \reporttablehead{Axis} & \reporttablehead{Variant} & \reporttablehead{Avg.} & \reporttablehead{$\Delta$}
        & \reporttablehead{Camera} & \reporttablehead{Robot} & \reporttablehead{Lang.} & \reporttablehead{Light}
        & \reporttablehead{BG.} & \reporttablehead{Noise} & \reporttablehead{Layout} & \reporttablehead{Total} & \reporttablehead{$\Delta$} \\
        \midrule

        \multirow{3}{*}{Teacher}
        & None
        & 91.9 & $-7.3$ & 8.4 & 8.1 & 33.6 & 51.5 & 34.7 & 9.9 & 33.3 & 24.0 & $-63.3$ \\
        & VGGT
        & 88.8 & $-10.4$ & 7.9 & 9.3 & 23.0 & 51.7 & 23.4 & 10.7 & 26.9 & 20.4 & $-66.9$ \\
        & DINOv2
        & 98.2 & $-1.0$ & 27.9 & 43.5 & 71.9 & 88.2 & 44.3 & 39.7 & 62.5 & 52.8 & $-34.5$ \\

        \midrule
        \multirow{3}{*}{State capacity}
        & $M=16$
        & 98.7 & $-0.5$ & 80.3 & 56.5 & 91.3 & 93.2 & 86.3 & 86.2 & 73.8 & 80.4 & $-6.9$ \\
        & $M=64$
        & 98.7 & $-0.5$ & 81.7 & 58.5 & \underline{95.7} & 89.3 & \underline{90.5} & \underline{93.5}
        & \underline{80.1} & \underline{83.7} & $-3.6$ \\
        & $M=128$
        & 98.3 & $-0.9$ & \underline{82.9} & \underline{59.1} & 91.5 & 92.6 & 83.3 & 87.9 & 76.9 & 81.6 & $-5.7$ \\

        \midrule
        \multirow{2}{*}{Extraction layer}
        & Layer 1
        & 72.9 & $-26.3$ & 30.2 & 21.5 & 52.5 & 47.2 & 12.5 & 28.2 & 41.0 & 33.6 & $-53.7$ \\
        & Layer 30
        & 98.6 & $-0.6$ & 74.2 & 54.3 & 90.4 & 89.3 & 75.4 & 83.0 & 73.1 & 76.7 & $-10.6$ \\

        \midrule
        \multirow{2}{*}{Future target}
        & Present only ($h_0$)
        & 95.2 & $-4.0$ & 41.0 & 34.8 & 53.1 & 75.6 & 45.0 & 45.3 & 54.4 & 49.0 & $-38.3$ \\
        & Near future only ($h_1$)
        & 98.5 & $-0.7$ & 63.8 & 46.0 & 87.4 & 89.6 & 74.9 & 74.9 & 75.8 & 72.4 & $-14.9$ \\

        \midrule
        State loss
        & $\Delta$-state only
        & 98.4 & $-0.8$ & 75.6 & 57.1 & 92.2 & \underline{95.6}
        & 77.6 & 87.7 & 79.9 & 80.4 & $-6.9$ \\
        & Absolute-state only
        & 98.0 & $-1.2$ & 73.3 & 56.5 & 90.3 & 85.8 & 78.8 & 75.3 & 68.6 & 74.9 & $-12.4$ \\

        \midrule
        Gradient
        & Detach video expert
        & 96.8 & $-2.4$ & 58.1 & 54.8 & 91.7 & 86.7 & 70.6 & 76.8 & 68.9 & 72.0 & $-15.3$ \\

        \midrule
        \reporttablehighlightrow
        \multicolumn{2}{l}{\textbf{Default ARPS (V-JEPA2)}}
        & \textbf{99.2} & --
        & \textbf{88.6} & \textbf{62.8} & \textbf{97.8} & \textbf{97.5}
        & \textbf{91.5} & \textbf{95.0} & \textbf{81.8} & \textbf{87.3} & -- \\

        \bottomrule
    \end{tabular}
    \arrayrulecolor{black}
    \endgroup
\end{table}

\textbf{Representation teacher.}
We first test the choice of representation space motivated by Sec.~\ref{sec:action_relevance}.
DINOv2 trails V-JEPA2 by only $1.0$ point on LIBERO, but by $34.5$ points on LIBERO-Plus.
Removing state supervision yields $24.0\%$ OOD success, and VGGT reaches $20.4\%$, below even the unsupervised-state variant.
These comparisons hold the state architecture and action objective fixed, demonstrating that the bottleneck alone does not explain the gains.
The ordering is consistent with the action-readability diagnostic, but V-JEPA2 and DINOv2 have much closer probe scores than OOD success rates.
Linear readability therefore does not fully explain control robustness. Sec.~\ref{sec:state_evidence} examines nuisance sensitivity as a further distinction.

\textbf{Temporal targets.}
We test whether learning future-related information in the state benefits control.
With the same V-JEPA2 teacher, present-only supervision reaches $49.0\%$ on LIBERO-Plus, versus $87.3\%$ with the default future objective.
Near-future-only supervision reaches $72.4\%$, $14.9$ points below the two-horizon objective.
On LIBERO, the corresponding gaps are only $4.0$ and $0.7$ points.
Present-only retains video co-training, so the gains highlight the value of explicitly supervising the predictive content of the action-facing state.
The diagnostic in Sec.~\ref{sec:future_increment} identifies useful action cues in observed futures.
These ablations show the control value of training the state to anticipate future information from current observations, particularly under distribution shift.

\textbf{Absolute and temporal-difference supervision.}
Temporal-difference supervision alone reaches $80.4\%$ on LIBERO-Plus, exceeding absolute-only supervision at $74.9\%$.
The joint objective reaches $87.3\%$, improving over these variants by $6.9$ and $12.4$ points, respectively.
Both targets therefore contribute to the full model's OOD performance.
The stronger difference-only result supports emphasizing temporal variation, while the additional gain from absolute targets is consistent with retaining the scene context in which changes occur.

\textbf{State capacity.}
The capacity sweep is non-monotonic.
All tested values of $M$ remain within $0.9$ LIBERO points of the default, while their LIBERO-Plus totals are $80.4\%$, $83.7\%$, and $81.6\%$ for $M=16$, $64$, and $128$.
The default $M=32$ achieves the highest success rate at $87.3\%$.
These results support a compact state but do not suggest that increasing capacity alone improves robustness.

\textbf{Extraction depth.}
Layer 1 reaches only $33.6\%$ on LIBERO-Plus, whereas layer 30 recovers strong in-domain performance but reaches $76.7\%$ OOD, $10.6$ points below layer 15.
The intermediate layer therefore offers the best control performance among the tested depths while also enabling early termination.
One possible explanation is that later features retain more appearance detail for video denoising. The depth comparison alone does not establish this mechanism.

\textbf{Joint optimization.}
Detaching the state predictor input leaves video supervision unchanged but prevents the state and action losses from shaping upstream video features.
This variant loses $2.4$ points on LIBERO and $15.3$ points on LIBERO-Plus.
Allowing these gradients to reach the video expert therefore contributes to the OOD gains, beyond training the state predictor over features shaped only by the video loss.

\textbf{End-to-end inference efficiency.}
On a single NVIDIA H20, ARPS decodes a complete action chunk in 306\,ms (3.26 calls/s), compared with 353\,ms (2.83 calls/s) for the official Fast-WAM implementation under identical settings, reducing end-to-end latency by $13\%$.
Early termination and the compact state reduce video-expert and action-denoising computation. App.~\ref{app:efficiency} reports the timing measurements.

\subsection{Does ARPS Learn a Useful Predictive State?}
\label{sec:state_evidence}

The ablations show that future supervision improves control, but do not characterize the learned state.
We examine its relationship to future behavior and its stability under visual changes, then test its role in closed-loop execution.

\begin{figure}[t]
    \centering
    \includegraphics[width=\linewidth]{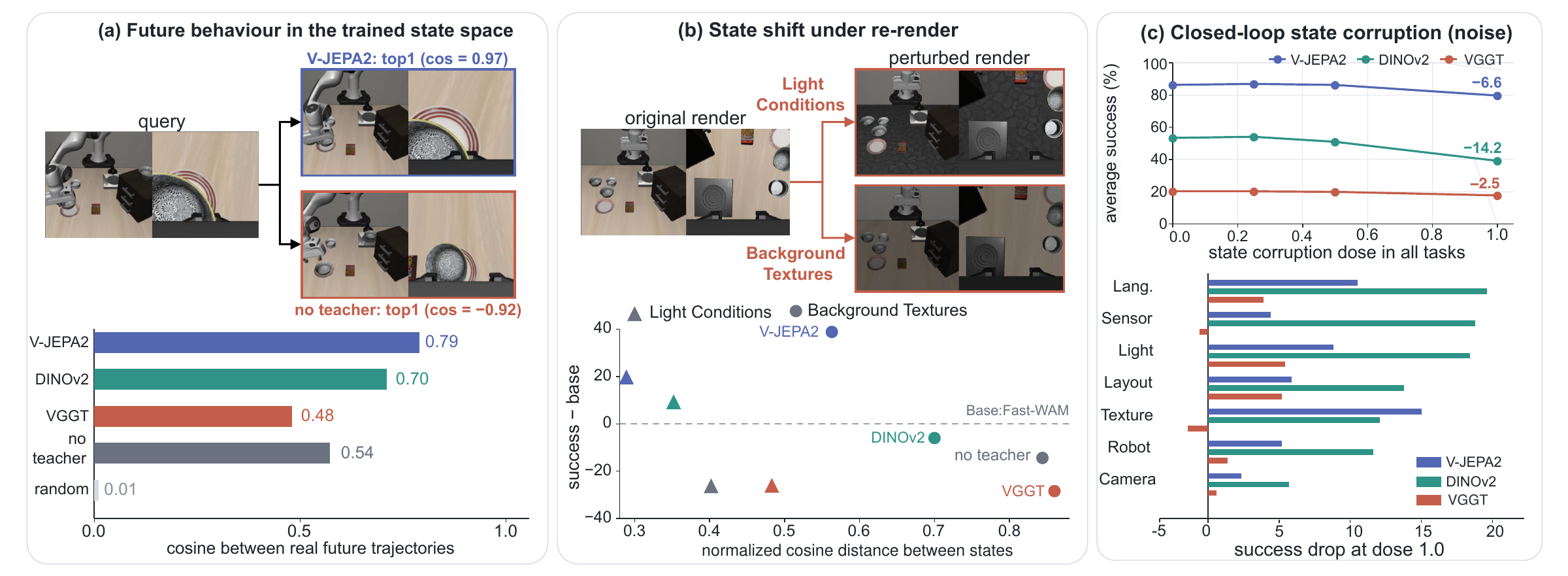}
    \caption{Post-training state analysis. (a) Future-trajectory similarity among state-space neighbors. (b) Normalized cosine distance before and after re-rendering, paired with success gain over Fast-WAM for the corresponding perturbation. (c) Closed-loop success under increasing state noise and per-category drops at the largest dose.}
    \label{fig:state_evidence}
\end{figure}

Fig.~\ref{fig:state_evidence}(a) tests whether nearby states correspond to similar future behavior.
Each of 256 query states retrieves its nearest neighbor by state cosine similarity, without using future trajectories or decoding heads.
We then compare the pair's recorded 32-step end-effector trajectories.
Mean trajectory cosine similarities are $0.79$, $0.70$, $0.48$, and $0.54$ for V-JEPA2, DINOv2, VGGT, and None, respectively, versus $0.01$ for random neighbors.
V-JEPA2 and DINOv2 both exceed the variant without a representation teacher.
The latter also exceeds random retrieval, so teacher supervision is not the only source of this structure.

Behavioral similarity alone does not establish robustness to appearance changes.
In Fig.~\ref{fig:state_evidence}(b), we re-render the same simulator state under different lighting or table textures and measure the cosine distance between the resulting states, normalized by the average distance between different scenes.
We pair this distance with the LIBERO-Plus success gain over Fast-WAM for the corresponding perturbation category.
V-JEPA2 has the smallest state shift and the largest success gain under both perturbations.
DINOv2 retains substantial future-behavior similarity in Fig.~\ref{fig:state_evidence}(a), yet its larger shifts under re-rendering accompany smaller control gains.
These results associate state stability with robustness to the tested visual changes, without establishing a causal relationship.

We next perturb the state supplied to the action expert with Gaussian noise scaled by the state root-mean-square magnitude.
Fig.~\ref{fig:state_evidence}(c) reports results on the same 1,000 LIBERO-Plus tasks at every dose.
At the largest dose, success drops by $6.6$, $14.2$, and $2.5$ percentage points for V-JEPA2, DINOv2, and VGGT, respectively.
The declines show that perturbing the deployed state changes closed-loop performance.
VGGT's low initial success rate limits the scope for a further decline, so the drop alone cannot rank state usefulness across variants.
This intervention tests the state as a whole and does not isolate its future-related content.
The ablations show the control benefit of future supervision, while these analyses characterize the resulting state and verify its use during policy execution.

\section{Conclusion}
\label{sec:conclusion}

We studied how a generation-free WAM should construct its action-facing state.
Our diagnostics show that predictable features need not be action-readable, useful information is spatially concentrated, and future observations provide complementary action cues.
ARPS turns these findings into a compact, horizon-conditioned state supervised by absolute and temporal-difference targets from a frozen video teacher while using only current observations at inference.
Across three benchmarks, ARPS delivers strong control, especially under distribution shift, while ablations and state analyses support the roles of teacher representation, future supervision, and state construction.
Overall, the results highlight the value of explicitly designing the predictive information delivered to the policy.

\textbf{Limitations.}
The substantial computational cost of WAM pretraining makes it difficult to evaluate additional backbones, so our study covers only one.
All experiments are currently conducted in simulation, and we plan to evaluate ARPS on real robots in future work.

\begingroup
\setlength{\bibsep}{4pt}
\bibliographystyle{plainnat}
\bibliography{references}
\endgroup
\clearpage
\appendix

\section{Additional Implementation Details}
\label{app:implementation}

LIBERO observations contain two camera views at a combined resolution of $224\times448$, and the policy predicts a 32-step action chunk.
At inference, ActionDiT uses 10 flow-matching denoising steps.
Each training clip contains 33 raw video frames, from which we sample nine frames $s_0,\ldots,s_8$.
The video VAE is temporally causal with a compression factor of four and maps these frames to three latent times: the current latent corresponds to $s_0$, while future latents $h=1$ and $h=2$ correspond to $s_{1:4}$ and $s_{5:8}$, respectively.
The two supervised horizons are latent indices $\mathcal H=\{1,2\}$, each covering 16 raw steps.

For each sampled endpoint $u$, the frozen V-JEPA2 teacher encodes a causal four-frame clip containing frames at $u-12$, $u-8$, $u-4$, and $u$.
The resulting representation therefore summarizes recent history but contains no frame after $u$.
In particular, the present-only target is causal, although it is not a single-frame feature.
At the beginning of a sequence, unavailable indices are clamped to frame index zero, which repeats the first frame.
We average the output tokens along the tubelet-time axis to obtain a $P=128$, $D=1024$ spatial field and assign that field to endpoint $u$.
Current and future fields use the same clip construction, encoder, temporal pooling, and spatial layout, differing only in their endpoint.
Because clips at nearby endpoints can overlap, short-horizon differences represent temporally smoothed changes rather than instantaneous frame differences.
To match the VAE timeline, we use the field assigned to $s_0$ as $T_t$ and average the four fields assigned to $s_{1:4}$ and $s_{5:8}$ to form $T_{t+1}$ and $T_{t+2}$.
Before caching, we apply per-channel z-score standardization using statistics computed from the training-set teacher features.
This is diagonal standardization rather than full-covariance whitening.
All cosine and smooth-$\ell_1$ losses and ridge analyses operate in this space, so their geometry, channel weighting, and regularization are defined after standardization.
The predictive state has width 1024, with 16 states assigned to each horizon.
Each horizon uses independent absolute-representation and temporal-difference position-query decode heads in the frozen V-JEPA2 space.
State losses use the valid-frame mask and average the cosine plus smooth-$\ell_1$ matching loss over horizons and all 128 spatial positions.
The complete 32-state representation is projected once and appended to the shared context consumed by every ActionDiT block.
During training, the video expert is flow-matching supervised on all three latent frames, while the action expert is flow-matching supervised on the 32-step action chunk with shift $5.0$.

\textbf{Evaluation protocol.}
For LIBERO, we run 50 evaluation rollouts per task.
For LIBERO-Plus, we follow its evaluation protocol over 10,030 perturbed task variants spanning seven shift categories, with one rollout per variant.
For RoboTwin 2.0, both the Clean and Random settings contain 50 tasks, and we run 100 evaluation rollouts per task in each setting (5,000 rollouts per setting).
The Fast-WAM results in Tabs.~\ref{tab:robotwin_results} and~\ref{tab:libero_plus} are from our own reproduction runs under the evaluation protocols above.
The Cosmos-Policy results on LIBERO-Plus are taken from the evaluation by \citet{WAMRobustness}, while the Faster-WAM results are reported by \citet{FasterWAM}.

For LIBERO, we train for approximately 22k steps with an effective batch size of 128 on 8 H20 GPUs.
For RoboTwin 2.0, we train for approximately 30k steps with an effective batch size of 1024 on 32 H20 GPUs.
For both benchmarks, we use AdamW with $\beta=(0.9,0.95)$ and weight decay $0.01$.
The learning rate is $1\times10^{-4}$ with a cosine schedule and $5\%$ linear warmup.

\textbf{Temporal-target ablations.}
In the present-only variant, the decoding heads are trained to align directly with the current teacher representation $T_t$, replacing the future absolute and temporal-difference targets.
The near-future-only variant retains $M=32$ state tokens, all assigned to the single supervised horizon $h=1$.
Its action-facing state therefore has the same token count as the default two-horizon model.

\section{Per-Task RoboTwin 2.0 Results}
\label{app:robotwin_per_task}

Tab.~\ref{tab:robotwin_per_task} reports the success rate of ARPS on every RoboTwin 2.0 task.
Tasks are ordered alphabetically down the left block and then down the right block.
The averages exactly reproduce the aggregate results in Tab.~\ref{tab:robotwin_results}.

\begin{table}[H]
    \centering
    \begingroup
    \caption{Per-task success rates (\%) on RoboTwin 2.0. Each task is evaluated with 100 rollouts in both settings.}
    \label{tab:robotwin_per_task}
    \footnotesize
    \setlength{\tabcolsep}{4.5pt}
    \renewcommand{\arraystretch}{0.91}
    \arrayrulecolor{TableRule}
    \begin{tabular}{@{}lrr@{\hspace{18pt}}lrr@{}}
        \toprule
        \reporttablepanelrow
        \reporttablehead{Task} & \reporttablehead{Clean} & \reporttablehead{Random} & \reporttablehead{Task} & \reporttablehead{Clean} & \reporttablehead{Random} \\
        \midrule
        \texttt{adjust\_bottle} & 100.0 & 100.0 & \texttt{place\_can\_basket} & 68.0 & 73.0 \\
        \texttt{beat\_block\_hammer} & 100.0 & 95.0 & \texttt{place\_cans\_plasticbox} & 100.0 & 100.0 \\
        \texttt{blocks\_ranking\_rgb} & 97.0 & 98.0 & \texttt{place\_container\_plate} & 100.0 & 99.0 \\
        \texttt{blocks\_ranking\_size} & 85.0 & 79.0 & \texttt{place\_dual\_shoes} & 100.0 & 96.0 \\
        \texttt{click\_alarmclock} & 100.0 & 98.0 & \texttt{place\_empty\_cup} & 100.0 & 100.0 \\
        \texttt{click\_bell} & 100.0 & 100.0 & \texttt{place\_fan} & 91.0 & 92.0 \\
        \texttt{dump\_bin\_bigbin} & 100.0 & 96.0 & \texttt{place\_mouse\_pad} & 69.0 & 69.0 \\
        \texttt{grab\_roller} & 100.0 & 100.0 & \texttt{place\_object\_basket} & 92.0 & 87.0 \\
        \texttt{handover\_block} & 78.0 & 63.0 & \texttt{place\_object\_scale} & 87.0 & 88.0 \\
        \texttt{handover\_mic} & 100.0 & 100.0 & \texttt{place\_object\_stand} & 92.0 & 92.0 \\
        \texttt{hanging\_mug} & 55.0 & 48.0 & \texttt{place\_phone\_stand} & 96.0 & 98.0 \\
        \texttt{lift\_pot} & 100.0 & 100.0 & \texttt{place\_shoe} & 96.0 & 100.0 \\
        \texttt{move\_can\_pot} & 94.0 & 97.0 & \texttt{press\_stapler} & 100.0 & 95.0 \\
        \texttt{move\_pillbottle\_pad} & 94.0 & 93.0 & \texttt{put\_bottles\_dustbin} & 79.0 & 86.0 \\
        \texttt{move\_playingcard\_away} & 100.0 & 100.0 & \texttt{put\_object\_cabinet} & 88.0 & 82.0 \\
        \texttt{move\_stapler\_pad} & 77.0 & 75.0 & \texttt{rotate\_qrcode} & 87.0 & 89.0 \\
        \texttt{open\_laptop} & 100.0 & 100.0 & \texttt{scan\_object} & 86.0 & 83.0 \\
        \texttt{open\_microwave} & 96.0 & 76.0 & \texttt{shake\_bottle} & 100.0 & 100.0 \\
        \texttt{pick\_diverse\_bottles} & 82.0 & 77.0 & \texttt{shake\_bottle\_horizontally} & 100.0 & 100.0 \\
        \texttt{pick\_dual\_bottles} & 96.0 & 91.0 & \texttt{stack\_blocks\_three} & 94.0 & 93.0 \\
        \texttt{place\_a2b\_left} & 92.0 & 95.0 & \texttt{stack\_blocks\_two} & 100.0 & 100.0 \\
        \texttt{place\_a2b\_right} & 96.0 & 91.0 & \texttt{stack\_bowls\_three} & 89.0 & 85.0 \\
        \texttt{place\_bread\_basket} & 95.0 & 91.0 & \texttt{stack\_bowls\_two} & 96.0 & 100.0 \\
        \texttt{place\_bread\_skillet} & 92.0 & 94.0 & \texttt{stamp\_seal} & 64.0 & 75.0 \\
        \texttt{place\_burger\_fries} & 100.0 & 100.0 & \texttt{turn\_switch} & 63.0 & 66.0 \\
        \midrule
        \textbf{Mean over 50 tasks} & \textbf{91.32} & \textbf{90.10} & & & \\
        \bottomrule
    \end{tabular}
    \arrayrulecolor{black}
    \endgroup
\end{table}

ARPS succeeds on all 100 rollouts in both settings for 13 tasks, while the aggregate gap between Clean and Random is only $1.22$ points.
Performance nevertheless varies substantially by task.
The lowest success rates occur on \texttt{hanging\_mug}, \texttt{handover\_block}, and \texttt{turn\_switch}.
Randomization produces the largest declines on \texttt{open\_microwave} and \texttt{handover\_block}, whereas several tasks remain unchanged or improve slightly.
The aggregate score therefore reflects broad robustness rather than uniform robustness across manipulation skills.

\section{Full LIBERO Ablation Results}
\label{app:full_libero_ablation}

Tab.~\ref{tab:libero_ablation_full} expands the LIBERO average in Tab.~\ref{tab:arps_ablation} into the four evaluation suites.

\begin{table}[H]
    \centering
    \begingroup
    \caption{Complete LIBERO ablation results. $M$ is the total number of predictive states, and $\Delta$ is the average-rate change from default ARPS.}
    \label{tab:libero_ablation_full}
    \small
    \setlength{\tabcolsep}{5pt}
    \renewcommand{\arraystretch}{0.92}
    \arrayrulecolor{TableRule}
    \begin{tabular}{@{}llcccccc@{}}
        \toprule
        \reporttablepanelrow
        \reporttablehead{Axis} & \reporttablehead{Variant} & \reporttablehead{Spatial} & \reporttablehead{Object} & \reporttablehead{Goal} & \reporttablehead{Long} & \reporttablehead{Avg.} & \reporttablehead{$\Delta$} \\
        \midrule

        \reporttablehighlightrow

        \multicolumn{2}{l}{\textbf{Default ARPS (V-JEPA2)}}
        & \textbf{99.0} & \textbf{100.0} & \underline{98.6} & \textbf{99.0} & \textbf{99.2} & -- \\

        \midrule
        \multirow{3}{*}{Teacher}
        & None
        & 93.8 & 96.2 & 95.2 & 82.4 & 91.9 & $-7.3$ \\
        & VGGT
        & 93.0 & 96.6 & 95.4 & 70.4 & 88.8 & $-10.4$ \\
        & DINOv2
        & 98.6 & \underline{99.6} & 98.2 & 96.2 & 98.2 & $-1.0$ \\

        \midrule
        \multirow{3}{*}{State capacity}
        & $M=16$
        & 98.4 & \underline{99.8} & \underline{98.6} & \underline{98.0} & \underline{98.7} & $-0.5$ \\
        & $M=64$
        & \underline{98.8} & \textbf{100.0} & 98.2 & 97.8 & \underline{98.7} & $-0.5$ \\
        & $M=128$
        & 98.4 & \textbf{100.0} & 98.2 & 96.6 & 98.3 & $-0.9$ \\

        \midrule
        \multirow{2}{*}{Extraction layer}
        & Layer 1
        & 82.0 & 79.6 & 77.6 & 52.4 & 72.9 & $-26.3$ \\
        & Layer 30
        & 98.4 & \underline{99.8} & \textbf{99.0} & 97.2 & 98.6 & $-0.6$ \\

        \midrule
        \multirow{2}{*}{Future target}
        & Present only ($h_0$)
        & 95.6 & 97.0 & 94.0 & 94.2 & 95.2 & $-4.0$ \\
        & Near future only ($h_1$)
        & 98.0 & \textbf{100.0} & 98.4 & 97.4 & 98.5 & $-0.7$ \\

        \midrule
        State loss
        & $\Delta$-state only
        & \textbf{99.0} & 99.4 & 98.0 & 97.2 & 98.4 & $-0.8$ \\
        & Absolute-state only
        & 98.0 & 99.6 & 98.2 & 96.2 & 98.0 & $-1.2$ \\

        \midrule
        Gradient
        & Detach video expert
        & 97.4 & 98.4 & 96.6 & 94.8 & 96.8 & $-2.4$ \\

        \bottomrule
    \end{tabular}
    \arrayrulecolor{black}
    \endgroup
\end{table}

\textbf{Additional objective and interface ablations.}
We further test whether future-representation supervision can replace the video objective and whether an explicit state bottleneck is needed.
Tab.~\ref{tab:additional_design_ablation} reports these comparisons on LIBERO-Plus.
JEPA-dense replaces the learned state $Z_t$ with dense video-expert hidden states $H_t^\ell$ as the visual context supplied to the action expert and the input to the representation decoding heads.
It retains the same extraction layer, V-JEPA2 teacher, future horizons, absolute and temporal-difference supervision, video and action objectives, and training settings as ARPS.

\begin{table}[H]
    \centering
    \begingroup
    \caption{Additional design ablations on LIBERO-Plus. No video loss sets $\mathcal L_{\mathrm{video}}=0$. JEPA-dense replaces the ARPS state with dense video-expert hidden states while retaining the teacher and supervision objectives. $\Delta$ is the change from default ARPS.}
    \label{tab:additional_design_ablation}
    \small
    \setlength{\tabcolsep}{4.2pt}
    \renewcommand{\arraystretch}{1.15}
    \arrayrulecolor{TableRule}
    \begin{tabular}{@{}lccccccccc@{}}
        \toprule
        \reporttablepanelrow
        \reporttablehead{Variant} & \reporttablehead{Camera} & \reporttablehead{Robot} & \reporttablehead{Language} & \reporttablehead{Light} & \reporttablehead{Background} & \reporttablehead{Noise} & \reporttablehead{Layout} & \reporttablehead{Total} & \reporttablehead{$\Delta$} \\
        \midrule
        \reporttablehighlightrow
        \textbf{Default ARPS}
        & \textbf{88.6} & \textbf{62.8} & \textbf{97.8} & \textbf{97.5}
        & \textbf{91.5} & \textbf{95.0} & \textbf{81.8} & \textbf{87.3} & -- \\
        No video loss
        & 60.7 & 43.7 & 82.3 & 89.4 & 51.0 & 67.1 & 60.8 & 64.6 & $-22.7$ \\
        JEPA-dense
        & \underline{66.5} & \underline{59.9} & \underline{89.4} & \underline{91.8}
        & \underline{85.0} & \underline{76.7} & \underline{75.3} & \underline{76.8} & $-10.5$ \\
        \bottomrule
    \end{tabular}
    \arrayrulecolor{black}
    \endgroup
\end{table}

Removing the video objective reduces LIBERO-Plus success by $22.7$ points, indicating that video co-training provides complementary benefits beyond future-state and action supervision.
With the same V-JEPA2 teacher, JEPA-dense achieves $76.8\%$, remaining $10.5$ points below ARPS.
This comparison supports the compact state interface over the evaluated dense-alignment alternative.

\section{Diagnostic Details}
\label{app:diagnostics}

\textbf{Data and probe protocol.}
We select 2,048 samples at evenly spaced indices from the LIBERO training data.
We use a deterministic 60/20/20\% split at the episode level, so samples from one episode never cross splits.
Each sample contains the current teacher field, teacher fields at two future horizons, and a normalized 32-step action chunk with 224 scalar targets.
Teacher features come from the same per-channel standardized cache used to train ARPS.
The future fields follow the video VAE timeline, with each future latent matched to the mean teacher feature over its corresponding group of four sampled frames.

All probes use ridge regression.
The regularization coefficient is selected on validation data from a 17-point logarithmic grid spanning $10^{-4}$ to $10^4$ for the global teacher-to-action probe and $10^{-3}$ to $10^5$ for the spatial and future-information probes.
We report test $R^2=1-\mathrm{SSE}/\mathrm{SST}$ for each regression target.
For action-readability probes, the target is the normalized action chunk, and the training-set action mean defines the constant-prediction baseline in $\mathrm{SST}$.
For future-change predictability, the target is $\Delta T_{h_2}$. Prediction errors and the reference variation are evaluated in the teacher feature space, rather than the action space.
High-dimensional concatenated features are fit with the exact dual ridge solution so that joint probes are not disadvantaged by an undersized solver.
As a sanity check, shuffling the feature-to-action pairing collapses the probe score.

\textbf{Action readability and predictability.}
For global action readout, we spatially average each selected feature field before fitting the action probe.
The camera-aware variant averages the two $8$-column camera regions separately and concatenates their features.
Fig.~\ref{fig:diagnosis}(a) visualizes position-wise action readability rather than PCA components.
For each teacher, we first project its features to 128 dimensions with PCA for a dimension-matched comparison, then fit an independent action probe at each of the 128 spatial locations.
The heatmap reports the resulting test $R^2$ at every location.
Predictability uses the same split and model-selection protocol but regresses from $T_t$ to the farther-horizon change $\Delta T_{h_2}$.

\textbf{Observed future information.}
The oracle diagnostic in Fig.~\ref{fig:diagnosis}(c) compares probes from $T_t$, $[T_t,\Delta T_{h_1}]$, and $[T_t,\Delta T_{h_1},\Delta T_{h_2}]$.
Its reported gain is the joint test $R^2$ minus the present-only test $R^2$.
Observed future fields are used only in this diagnostic and never as ARPS inputs at inference.

\textbf{Spatial concentration.}
Before ranking locations, we subtract the training mean from every current-frame feature.
An independent validation probe ranks the 128 locations by action $R^2$.
For each $m\in\{1,2,4,8,16,32,64,128\}$, we flatten the top-$m$ locations, refit ridge regression, and evaluate once on the test split.
The ranking never uses test data, and the full 128-location probe provides the reference readout.

\textbf{Post-training state analyses.}
For Fig.~\ref{fig:state_evidence}(a), we flatten and $\ell_2$-normalize the state tokens of each of 256 evaluation samples.
Within each teacher variant, every query retrieves the most similar other sample by cosine similarity.
We construct the recorded future behavior of each sample by cumulatively summing the end-effector position increments in its 32-step action chunk, then flatten and normalize the resulting trajectory.
The reported score averages the trajectory cosine similarity between each query and its retrieved neighbor.
Uniformly sampled neighbors provide the random baseline.
Neither teacher targets nor state decoding heads are used for retrieval.

For Fig.~\ref{fig:state_evidence}(b), clean and perturbed observations share the same simulator state and differ only in lighting or table texture.
We extract their predictive states and normalize their cosine distance by the corresponding variant's average cosine distance between different scenes.
The vertical axis reports the success-rate change over Fast-WAM on the matching LIBERO-Plus perturbation category, so zero denotes the Fast-WAM baseline.

For Fig.~\ref{fig:state_evidence}(c), we add isotropic Gaussian noise to the state supplied to the action expert at inference.
The noise standard deviation is $0$, $0.25$, $0.5$, or $1.0$ times the state root-mean-square magnitude.
Every dose is evaluated on the same stratified set of 1,000 LIBERO-Plus tasks with one rollout per task.
All model inputs and inference settings are otherwise unchanged, isolating the behavioral effect of perturbing the state values.

\section{State Invariance Across Perturbations}
\label{app:state_invariance}

\begin{figure}[H]
    \centering
    \includegraphics[width=0.9\textwidth]{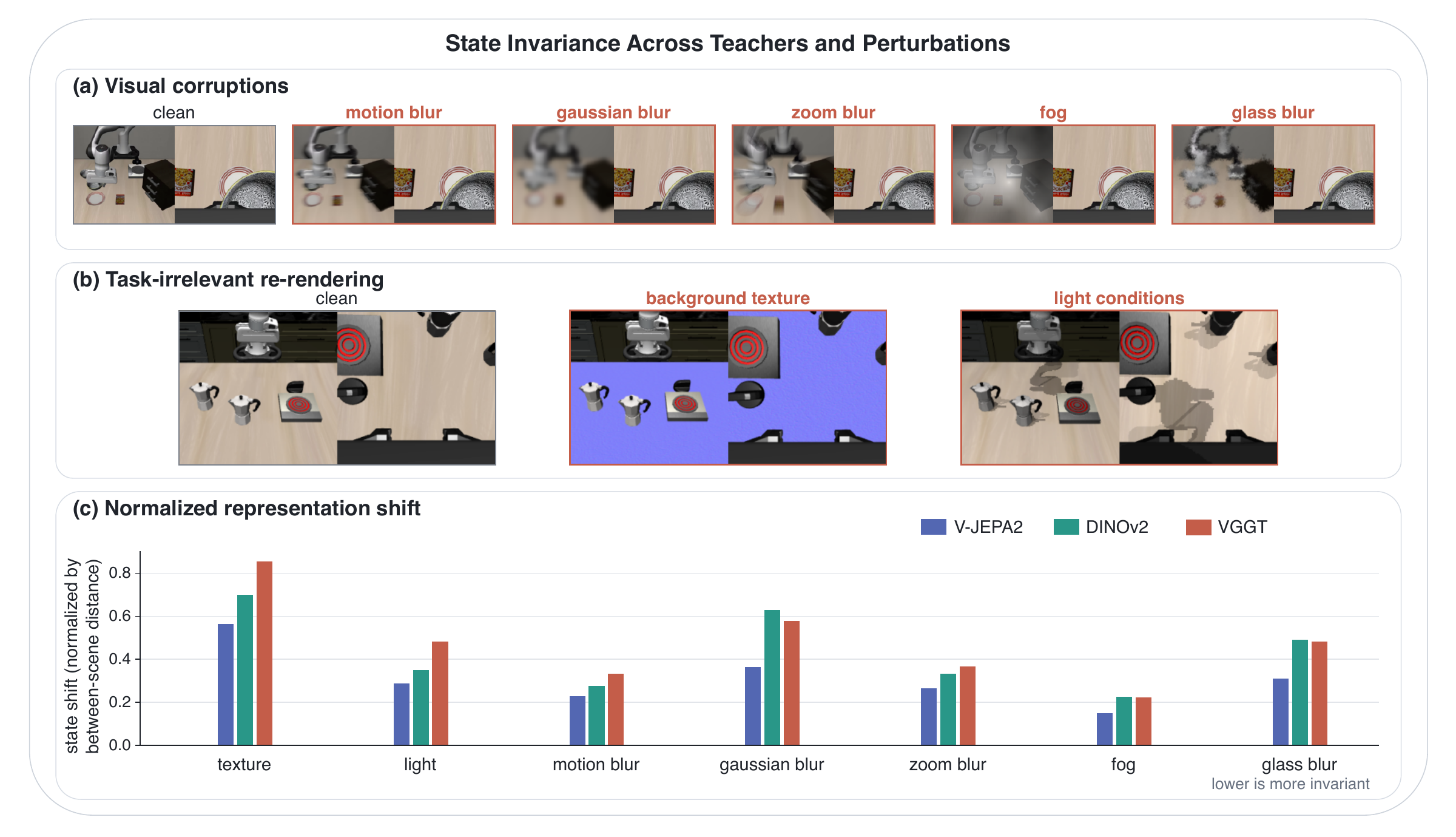}
    \caption{Predictive-state invariance across visual perturbations. (a) Examples of image-space corruptions applied to a fixed observation. (b) Examples of task-irrelevant background and lighting changes produced by re-rendering the same simulator state. (c) Cosine distance between states from clean and perturbed observations, normalized by the average cosine distance between different scenes. Lower values indicate greater invariance.}
    \label{fig:state_invariance}
\end{figure}

Fig.~\ref{fig:state_invariance} extends the invariance analysis in Fig.~\ref{fig:state_evidence}(b) beyond background texture and lighting.
We apply five image-space corruptions to the same observation and also re-render a fixed simulator state with different background textures or lighting conditions.
For each clean and perturbed pair, we measure the cosine distance between their predictive states and normalize it by the average cosine distance between different scenes.
This normalization makes the scale comparable across teacher variants, with lower values indicating greater invariance to a task-irrelevant visual change.

V-JEPA2 produces the smallest normalized state shift under all seven perturbations.
The separation is especially clear for texture changes and Gaussian blur, where DINOv2 and VGGT move substantially farther from the clean state even though the underlying task is unchanged.
This broader comparison supports the association in Fig.~\ref{fig:state_evidence}(b): the teacher affects how strongly the learned state follows nuisance appearance, and V-JEPA2 yields the most stable interface among the evaluated representations.

\section{Inference Efficiency}
\label{app:efficiency}

We measure end-to-end action decoding on one NVIDIA H20 with the standard two-camera LIBERO input, a 32-step action horizon, and 10 action-denoising steps.
Both models remain resident on the same GPU and are measured in alternating runs after warmup.
Tab.~\ref{tab:efficiency_breakdown} reports the default 32-state ARPS because the 64-state variant has essentially identical latency.

\begin{table}[H]
    \centering
    \begingroup
    \caption{End-to-end latency and selected component timings on LIBERO.}
    \label{tab:efficiency_breakdown}
    \small
    \setlength{\tabcolsep}{6pt}
    \renewcommand{\arraystretch}{1.15}
    \arrayrulecolor{TableRule}
    \begin{tabular}{lcc}
        \toprule
        \reporttablepanelrow
        & \reporttablehead{Fast-WAM} & \reporttablehead{ARPS (32 states)} \\
        \midrule
        End-to-end latency (ms/call) & 353 & 306 \\
        Action-chunk generation rate (calls/s) & 2.83 & 3.26 \\
        VAE input encoding (ms) & 11.0 & 10.9 \\
        Video-expert prefill (ms) & 29.5 (30 layers) & 15.3 (15 layers) \\
        State predictor (ms) & N/A & 1.8 \\
        Action denoising, 10 steps (ms) & 29.9 & 27.0 \\
        \bottomrule
    \end{tabular}
    \arrayrulecolor{black}
    \endgroup
\end{table}

ARPS reduces end-to-end latency by $13\%$ under this measurement setup.
Early termination reduces video-expert prefill time, while the compact state adds a small prediction overhead and shortens action denoising.
The component measurements isolate selected GPU stages and are not an additive decomposition of end-to-end latency, so they do not quantify each stage's contribution to the total reduction.
The reported generation rate is the reciprocal of action-chunk latency. It does not denote the robot's low-level action execution frequency.

\section{Qualitative Failure Modes}
\label{app:failure_modes}

\begin{figure}[H]
    \centering
    \includegraphics[width=\textwidth]{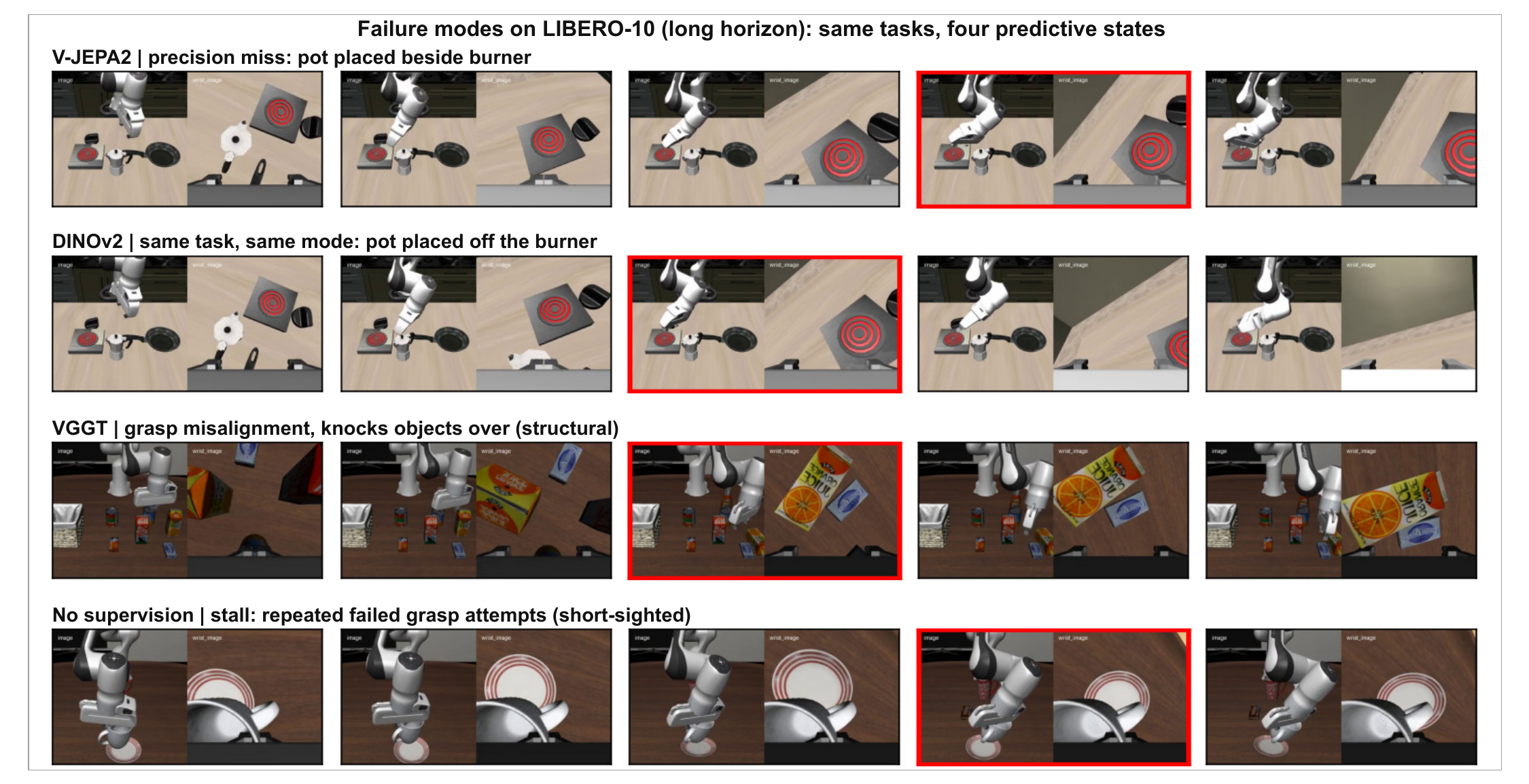}
    \caption{Representative failures on LIBERO-10. Red boxes mark the decisive errors: misplaced pots for V-JEPA2 and DINOv2, a grasp collision for VGGT, and repeated failed grasps for None.}
    \label{fig:failure_modes}
\end{figure}

Fig.~\ref{fig:failure_modes} shows representative failures on LIBERO-10 under different predictive-state supervision choices.
V-JEPA2 and DINOv2 preserve task progress but miss the final placement, while VGGT causes a grasp collision and the unsupervised variant stalls after repeated attempts.
These examples complement the aggregate ablation in Tab.~\ref{tab:arps_ablation}, but do not estimate failure frequency.

\end{document}